\pdfoutput=1

\documentclass[11pt]{article}

\usepackage[preprint]{acl}

\usepackage{times}
\usepackage{latexsym}

\usepackage[T1]{fontenc}

\usepackage[utf8]{inputenc}

\usepackage{microtype}

\usepackage{inconsolata}

\usepackage{graphicx}
\usepackage{subcaption}
\usepackage{float}
\usepackage{amsmath}
\usepackage{algorithm}
\usepackage{algpseudocode}
\usepackage{xcolor}
\usepackage{multirow}

\usepackage{changes}

\title{Cerberus: Efficient Inference with Adaptive Parallel Decoding and Sequential Knowledge Enhancement}

\author{Yuxuan Liu\textsuperscript{1,2},   Wenyuan Li\textsuperscript{1,2},  Laizhong Cui\textsuperscript{1,2}, Hailiang Yang\textsuperscript{1} \\
         \textsuperscript{1}Guangdong Laboratory of Artificial Intelligence and Digital Economy (SZ) \\  \textsuperscript{2}Shenzhen University \\  \textsuperscript {} 
          \{liuyuxuan, yanghailiang\}@gml.ac.cn}
\nolinenumbers

\begin{document}
\maketitle
\begin{abstract}
Large language models (LLMs) often face a bottleneck in inference speed due to their reliance on auto-regressive decoding. 
Recently, parallel decoding has shown significant promise in enhancing inference efficiency.
However, we have identified two key issues with existing parallel decoding frameworks: (1) decoding heads fail to balance prediction accuracy and the parallelism of execution, and
(2) parallel decoding is not a universal solution, as it can bring unnecessary overheads at some challenging decoding steps. 
To address these issues, we propose Cerberus,
an adaptive parallel decoding framework introduces the gating mechanism to enable the LLMs to adaptively choose appropriate decoding approaches at each decoding step, along with introducing a new paradigm of decoding heads that introduce the sequential knowledge while maintaining execution parallelism.
The experiment results demonstrate that the Cerberus can achieve up to 2.12x speed up compared to auto-regressive decoding, and outperforms one of the leading parallel decoding frameworks, Medusa, with a 10\% - 30\% increase in acceleration and superior generation quality.
\end{abstract}

\section{Introduction}

Generative large language models (LLMs), such as the GPT-4 \citep{achiam2023gpt}, LLaMA \citep{touvron2023llama}, PaLM \citep{chowdhery2023palm}, have demonstrated remarkable performance across various downstream applications. Unfortunately, these LLMs are constrained by auto-regressive decoding \citep{vaswani2017attention}, which limits them to generating one token per decoding step, making inference time-consuming.

Recently, several approaches based on parallel decoding, a type of speculative decoding algorithm, have been proposed to address this issue (\citealp{cai2024medusa}; \citealp{fu2024break}; \citealp{li2024eagle}; \citealp{he2024rest}; \citealp{ankner2024hydra}; \citealp{zhang2024recurrent}). The parallel decoding is a Draft-then-Verify decoding paradigm \citep{xia2024unlocking}, it introduces several decoding heads aligned with the original model head to simultaneously generate multiple token candidates. These candidates are then verified using a tree attention-based process, referred to as tree verification (\citealp{chen2024sequoia}; \citealp{Miao_2024}), allowing the LLM to generate multiple tokens at each decoding step.

\begin{figure*}[t]
  \begin{subfigure}{0.5\linewidth}
    \includegraphics[width=\linewidth]{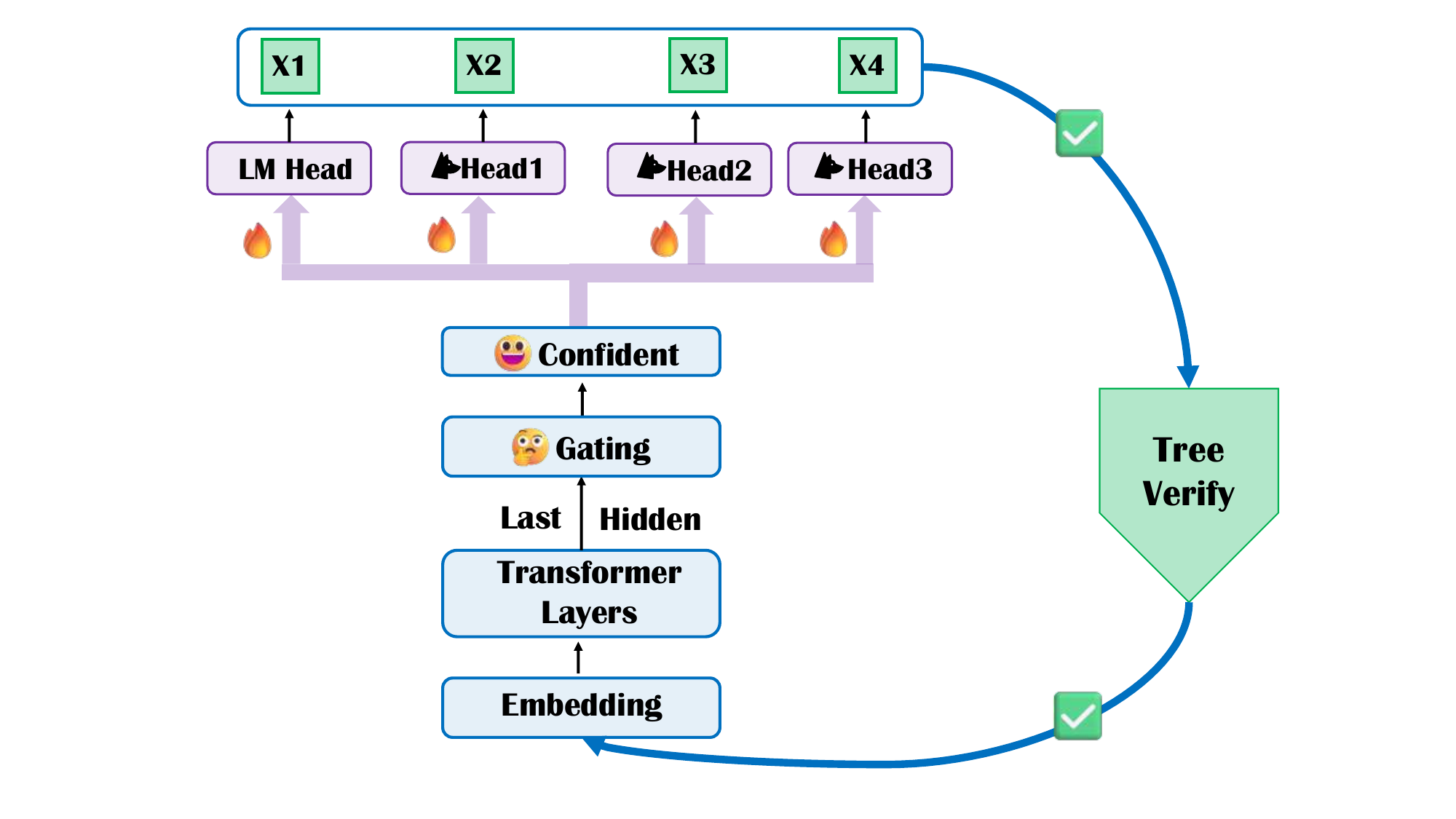}
    \caption{The inference flow when the LLM is confident}
    \label{fig:sub1}
  \end{subfigure}\hfill
  \begin{subfigure}{0.5\linewidth}
    \includegraphics[width=\linewidth]{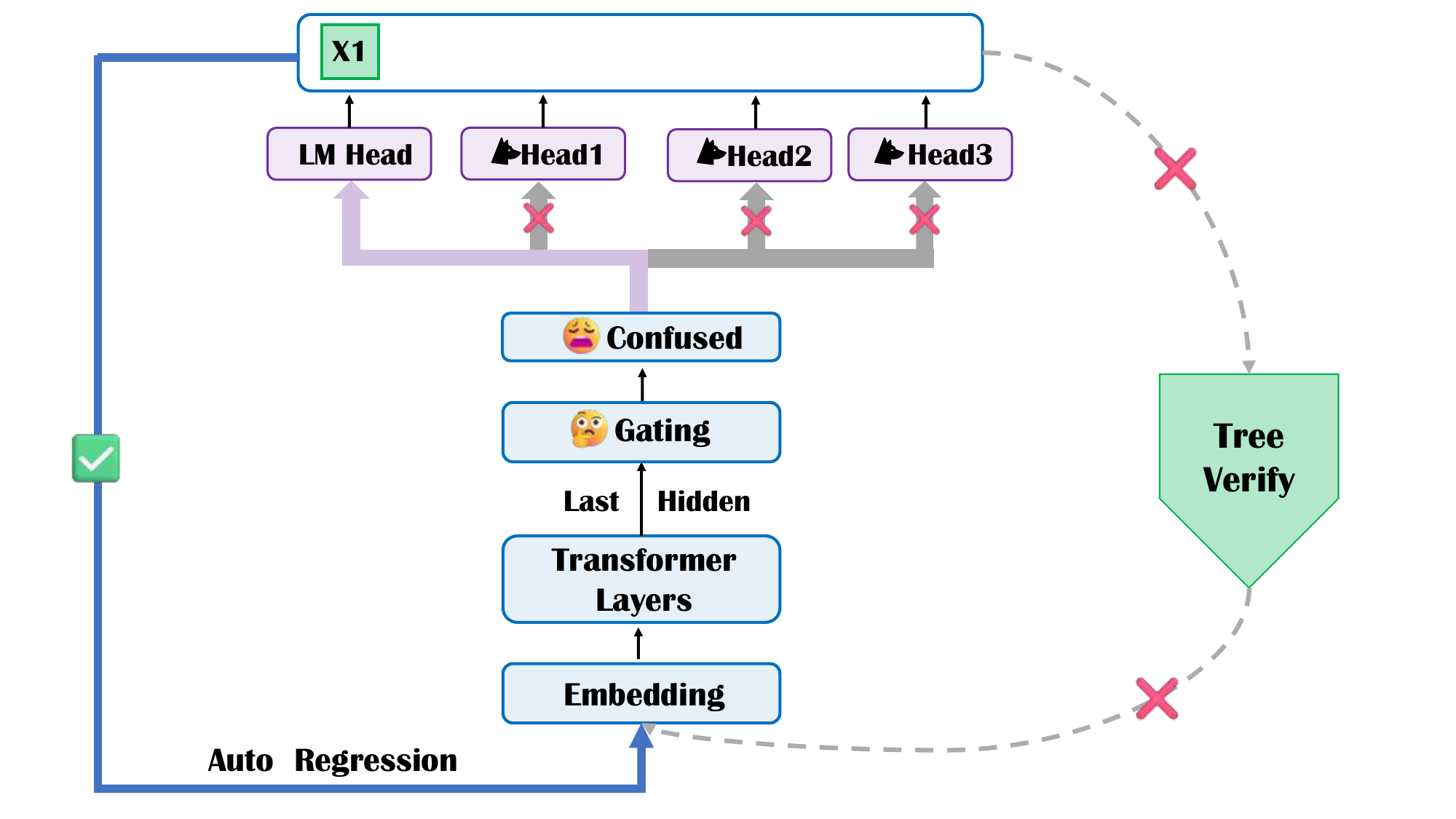}
    \caption{The inference flow when the LLM is confused}
    \label{fig:sub2}
  \end{subfigure}
  \caption{The implementation overview of Cerberus under two circumstances.}
\end{figure*}

However, we still observe two key challenges in existing parallel decoding frameworks. Firstly, \textbf{existing paradigms of decoding heads fail to balance prediction accuracy with execution parallelism.} Existing frameworks either use sequentially independent decoding heads or decoding heads connected in series. The former leads to low prediction accuracy, while the latter necessitates sequential execution of each decoding head, undermining parallelism. Secondly, \textbf{parallel decoding is not a universal solution to the entire decoding process.} Implementing parallel decoding introduces additional overheads. And since we observed that using parallel decoding at some challenging decoding steps can lead to low prediction accuracy, resulting in speeds comparable to auto-regressive decoding (Section \ref{unncessary overhead}), it can be wasteful to employ parallel decoding at these decoding steps.

To address these issues, we propose the Cerberus, an effective parallel decoding framework that incorporates a novel decoding heads paradigm called Cerberus Heads and a gating mechanism for selecting different decoding approaches. Our framework not only improves the prediction accuracy without compromising the execution parallelism but also reduces the overheads during the decoding process, achieving efficient inference.

Specifically, the \textbf{Cerberus heads}, as illustrated in \ref{fig:sub22}, introduces a sequential connection between the internal modules of each decoding head. This allows each head to learn information from previous positions, capture longer contexts, and make more accurate predictions. Furthermore, this implementation enables each decoding head to execute decoding synchronously without waiting for others, achieving both prediction accuracy and execution parallelism. For the \textbf{gating mechanism}, based on one of our observations (Section \ref{entropy}): the entropy of the last hidden states can reflect the prediction accuracy, and help us assess the model's confidence level, what we introduce is an entropy-based gating mechanism. As shown in Figure \ref {fig:sub1}, if the entropy of the last hidden states is below a specified threshold, we consider the model confident in the current prediction and then employ parallel decoding. Otherwise, only auto-regressive decoding will be used (Figure \ref {fig:sub2}). This can enable the LLM to adaptively choose different decoding approaches, reducing unnecessary overheads during the decoding process.

We evaluate Cerberus on MT-Bench \citep{zheng2024judging} with various settings of tree verification. The results show that Cerberus performs well across all settings, achieving a speedup ratio of up \textbf{1.57x-2.12x} over auto-regressive decoding. Furthermore, it also surpasses Medusa \citep{cai2024medusa}, one of the top-performing parallel decoding frameworks, with an additional \textbf{10\%-30\%} acceleration and better generation quality.

To summarize, our contributions are as follows:

• We conduct extensive analysis to identify two challenges in existing parallel decoding frameworks: (1) the decoding heads fail to balance the prediction accuracy and execution parallelism, and (2) parallel decoding is not universally efficient for the entire decoding process, as it can be wasteful at certain decoding steps.

• We propose Cerberus, a new framework that incorporates two essential components: (1) a novel paradigm of decoding heads that integrates sequential knowledge to enhance prediction accuracy while maintaining execution parallelism and (2) an entropy-based gating mechanism that enables the LLM to choose the appropriate decoding approach during each decoding step adaptively.

• We evaluate the Cerberus and show it achieves up to 2.12x speedup over auto-regressive decoding, and outperforms the Medusa with up to 30\% additional acceleration and better generation quality. 

\section{Related Work}

Researchers have introduced various technologies to improve the efficiency of LLM inference, such as batch processing \citep{yu2022orca}, group attention \citep{ainslie2023gqa}, operator fusion \citep{zhao2022apollo}, offloading \citep{sheng2023flexgen}, distillation\citep{zhou2023distillspec}, quantization \citep{xiao2023smoothquant}, and sparsification (\citealp{liu2023deja}, \citealp{song2023powerinfer}).

Similar to our approach are frameworks based on speculative decoding (\citealp{xia-etal-2023-speculative}; \citealp{chen2023accelerating}; \citealp{leviathan2023fast}; \citealp{Miao_2024}). Speculative decoding utilizes the draft-then-verify mechanism to empower LLMs to generate multiple tokens in a single forward propagation process. Mainstream speculative decoding frameworks can be categorized into the following two types \citep{xia2024unlocking}.

\subsection{Independent Drafting}
Independent drafting requires two models, a target model which is the original LLM, and a draft model which is a small model with a similar structure to the target model. During inference, the draft model generates tokens over several future time steps based on the input sequence through auto-regressive decoding, and the target model determines the ultimate output through parallel verification. The SpecDec \citep{xia-etal-2023-speculative} introduces a simple transformer model based on the encoder-decoder architecture as the draft model. Other researchers (\citealp{leviathan2023fast,chen2023accelerating}) directly utilize existing pre-trained language models, and propose the Speculative Sampling method to enhance the verification pass rate. The Specinfer \citep{Miao_2024} employs the boost-tuning method to train multiple Small Speculative Models (SSMs) and introduce the tree attention to use the LLM as a token tree verifier, and then to verify the tokens generated by the SSMs in parallel.

\subsection{Self-Drafting}
Self-drafting only requires fully leveraging the power of the original target model, instead of obtaining a suitable draft model for the LLM.

The parallel decoding framework accelerates decoding by introducing several decoding heads to the target model. Blockwise \citep{stern2018blockwise} and Medusa \citep{cai2024medusa} introduce independent FFN heads (illustrated in Figure \ref{fig:sub11}), each accepting the hidden states from the last transformer layer of the LLM as input, predicting the top-$k$ \citep {fan2018hierarchical} token candidates of the backbone model in the future independently. The Hydra \citep{ankner2024hydra} utilizes a serial connection between the entire decoding heads to propose a paradigm of sequentially dependent decoding heads, hence increasing the prediction accuracy of the decoding heads. The Recurrent-Drafter \citep{zhang2024recurrent}  is similar to Hydra, but it uses a paradigm of recurrent neural network (RNN) \citep{mikolov2012context} to introduce sequential knowledge between decoding heads instead of using a simple serial connection. Moreover, to improve the training method of the multi-token prediction paradigm, researchers \citep{gloeckle2024better} recently introduced a new framework to achieve a better and faster model. Unlike our framework which achieves parallel decoding by fine-tuning existing models, this paradigm integrates multiple decoding heads as part of the LLM and trains from scratch, demonstrating great performance on many downstream tasks.

Besides frameworks that introduce additional decoding heads, there are still many researchers who attempt to achieve speculative decoding by only utilizing a single target model. The EAGLE \citep{li2024eagle} attempts to achieve auto-regression at the feature level by training an auto-regression head within the LLM. The REST \citep{he2024rest} generates draft tokens by retrieving the existing databases, instead of utilizing the draft model. The Lookahead decoding \citep{fu2024break}, Ouroboros \citep{zhao2024ouroboros} and CLLMS \citep{kou2024cllms} improve the Jacobi iteration method to accelerate decoding.

\section{Parallel Decoding is Not a Universal Solution}

In this section, we present two key observations that inspired the design of Cerberus based on in-depth analysis of parallel decoding. We first demonstrate that parallel decoding is unsuitable for some decoding steps in Section \ref{unncessary overhead}. Then, we illustrate that we can decide whether to implement parallel decoding by assessing the entropy of hidden states at each decoding step in Section \ref{entropy}. 

\subsection{Unnecessary Overheads During Parallel Decoding}
\label{unncessary overhead}

In this section, we present our first key observation by conducting several experiments: parallel decoding is not universally efficient for the entire decoding process. We find that Parallel decoding can bring unnecessary overheads at some challenging decoding steps.

In parallel decoding, several decoding heads are added on top of the original LLM, and multiple token candidates are generated in parallel at each decoding step. All these token candidates are then verified by tree verification to determine the final tokens to be accepted.

\begin{figure}[H]
  \includegraphics[width=\columnwidth]{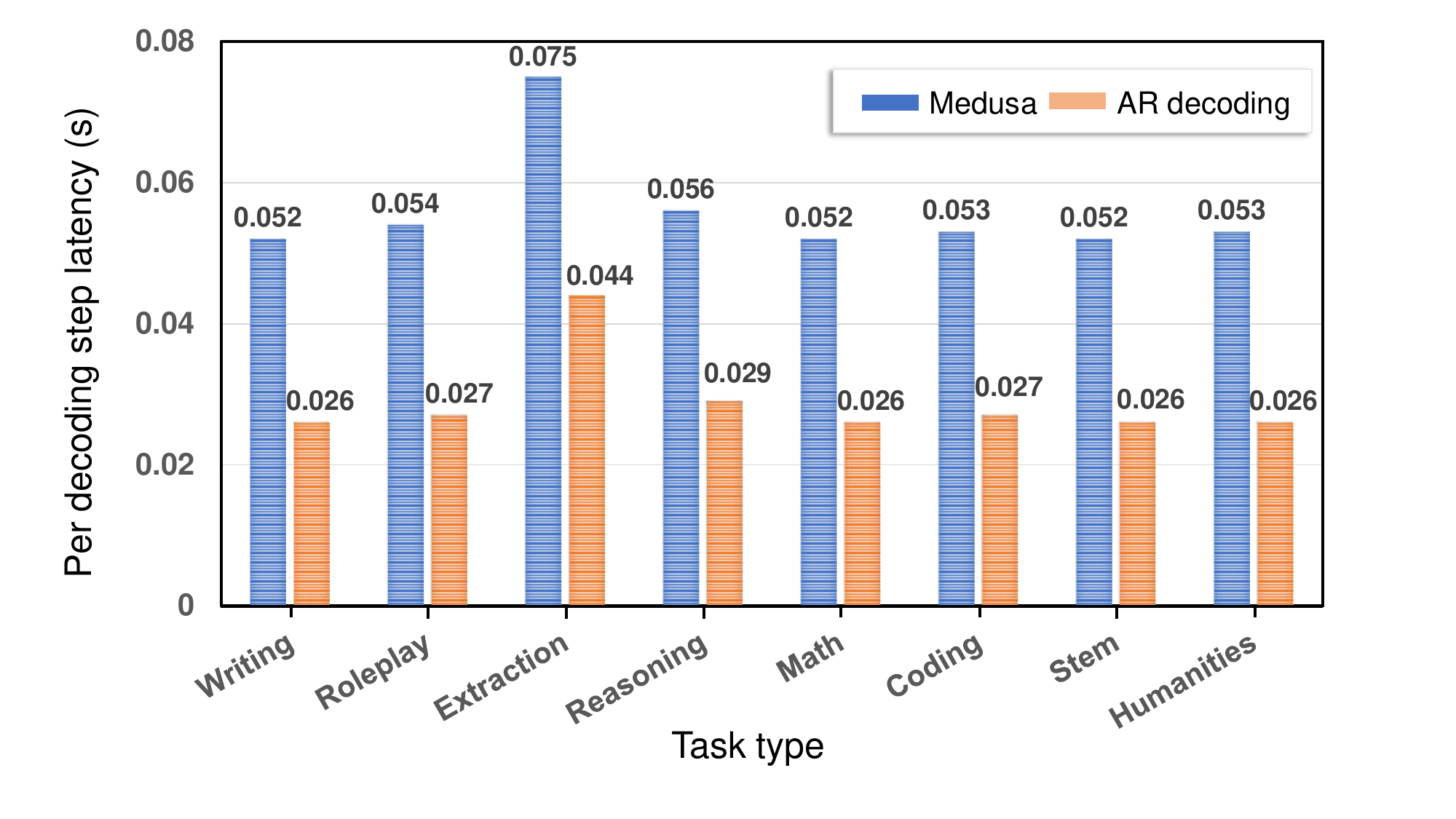}
  \caption{The average inference latency per decoding step. Since Medusa requires additional operations compared to auto-regressive decoding, it will bring more time spent.}
  \label{fig:time overhead}
\end{figure} 

Compared to auto-regressive decoding, parallel decoding can bring various additional overheads during inference, such as the time cost of using tree verification to verify token candidates, the computation resource requirements brought by decoding heads, and more complicated hardware implementations. As demonstrated in Figure \ref{fig:time overhead}, taking the Medusa \citep{cai2024medusa} as an example, Medusa’s average time spent on each decoding step is higher than auto-regressive decoding in all tasks of the MT-Bench \citep{zheng2024judging} due to the existence of the overheads. 

These overheads can sometimes lead to considerable accelerations, but sometimes they are just unnecessary. For the LLM, a complete task typically comprises both simple sub-tasks and challenging sub-tasks, corresponding to easy and hard decoding steps, respectively. For these easy decoding steps, parallel decoding can achieve high acceleration efficiency since it can attain high prediction accuracy to generate multiple tokens at each decoding step successfully. 

\begin{table}[H]
  \centering
  \begin{tabular}{c|c}
    \hline
    \multirow{2}{*}{\textbf{Task}} & \multirow{2}{*}{\textbf{Proportion}} \\
    & \\
    \hline
    \textbf{Avg.} &18.38\% \\
    \hline
    \textbf{Writing} &21.26\% \\
    \hline
    \textbf{Roleplay} &17.68\%\\
    \hline
    \textbf{Extraction} &23.52\%\\
    \hline
    \textbf{Reasoning} &12.14\% \\
    \hline
    \textbf{Math} &10.68\%\\
    \hline
    \textbf{Coding} &18.91\%\\
    \hline
    \textbf{Stem} &17.99\%\\
    \hline
    \textbf{Humanities} &24.84\%\\
    \hline
  \end{tabular}
  \caption{
   The proportion of decoding steps where no tokens generated by decoding heads are accepted during the entire Medusa decoding process, tested on MT-Bench.}
  \label{no tokens passed}
\end{table}

However, the performance of parallel decoding is not very good at those hard decoding steps that confuse the LLM. As shown in Table \ref{no tokens passed}, under tested on the eight types of downstream tasks of MT-Bench, an average of 18.38\% of parallel decoding steps fail to generate any additional token that passes the verification. In the tasks of Extraction and Humanities, this proportion can even reach 23.52\% and 24.84\%. These decoding steps without generating any token that passes the tree verification correspond to the situation that the model has low prediction accuracy, and the task being processed is quite puzzling for the model. At these challenging decoding steps, implementing parallel decoding is equivalent to consuming a lot of additional resources while achieving almost the same effect as auto-regressive decoding, which is a huge waste. Consequently, the overheads introduced by parallel decoding to accelerate inference do not achieve the expected effect, rendering it not worthwhile and unnecessary in this case.

\subsection{Entropy of Last Hidden States Can Reflect the Prediction Accuracy }
\label{entropy}
In the previous section, we demonstrated that parallel decoding is efficient only at decoding steps with high prediction accuracy, otherwise it will bring unnecessary overheads. In this section, we figure out how we determine whether to implement parallel decoding during each decoding step.

\begin{figure}[t]
  \includegraphics[width=\columnwidth]{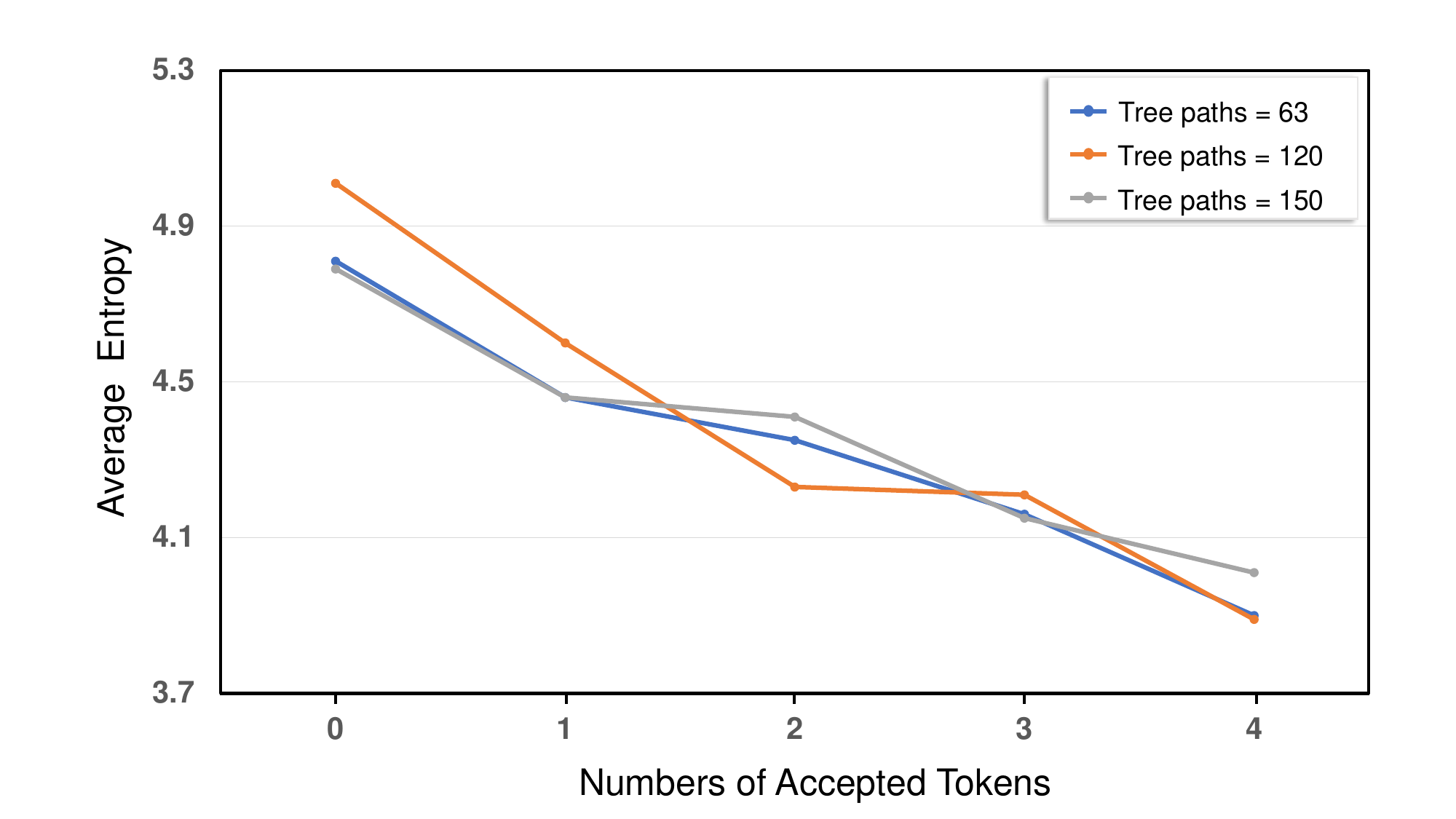}
  \caption{The average entropy of the last hidden states when accepting different numbers of tokens, with setting 4 decoding heads. We conduct this experiment under three different tree paths, the setting of the tree path is a crucial component of tree verification, detailed presentation can be seen in Section \ref{experimental setup}.}
  \label{fig:tokens entropy}
\end{figure}

In Figure \ref{fig:tokens entropy}, we measure the relationship between the entropy of the last hidden states and the number of accepted tokens per decoding step using three different tree settings. The result shows there is a negative correlation between the entropy of the last hidden states and the number of accepted tokens. And since more accepted tokens are always accompanied by higher prediction accuracy and a higher model confidence level, we can obtain the model's confidence level by simply assessing the entropy of the last hidden states. Therefore, this entropy can be a criterion to determine whether the current decoding step is suitable for parallel decoding.

Based on this analysis, we propose our second key observation: the entropy of the last hidden states can reflect the prediction accuracy and help assess the model’s confidence level.

\section{Cerberus}

\begin{figure*}[t]
  \begin{subfigure}{0.5\linewidth}
    \includegraphics[width=\linewidth]{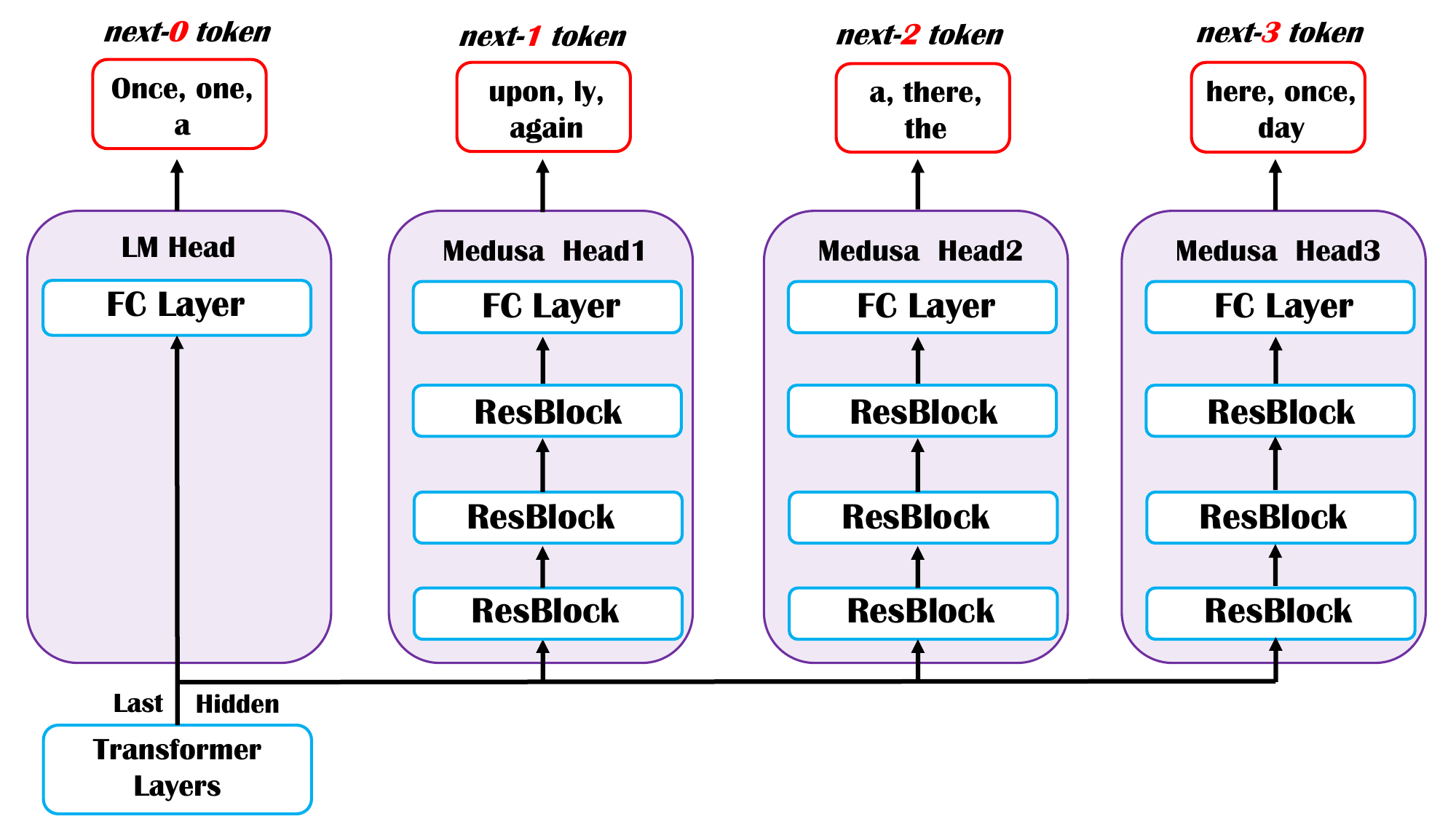}
    \caption{Medusa heads}
    \label{fig:sub11}
  \end{subfigure}\hfill
  \begin{subfigure}{0.5\linewidth}
    \includegraphics[width=\linewidth]{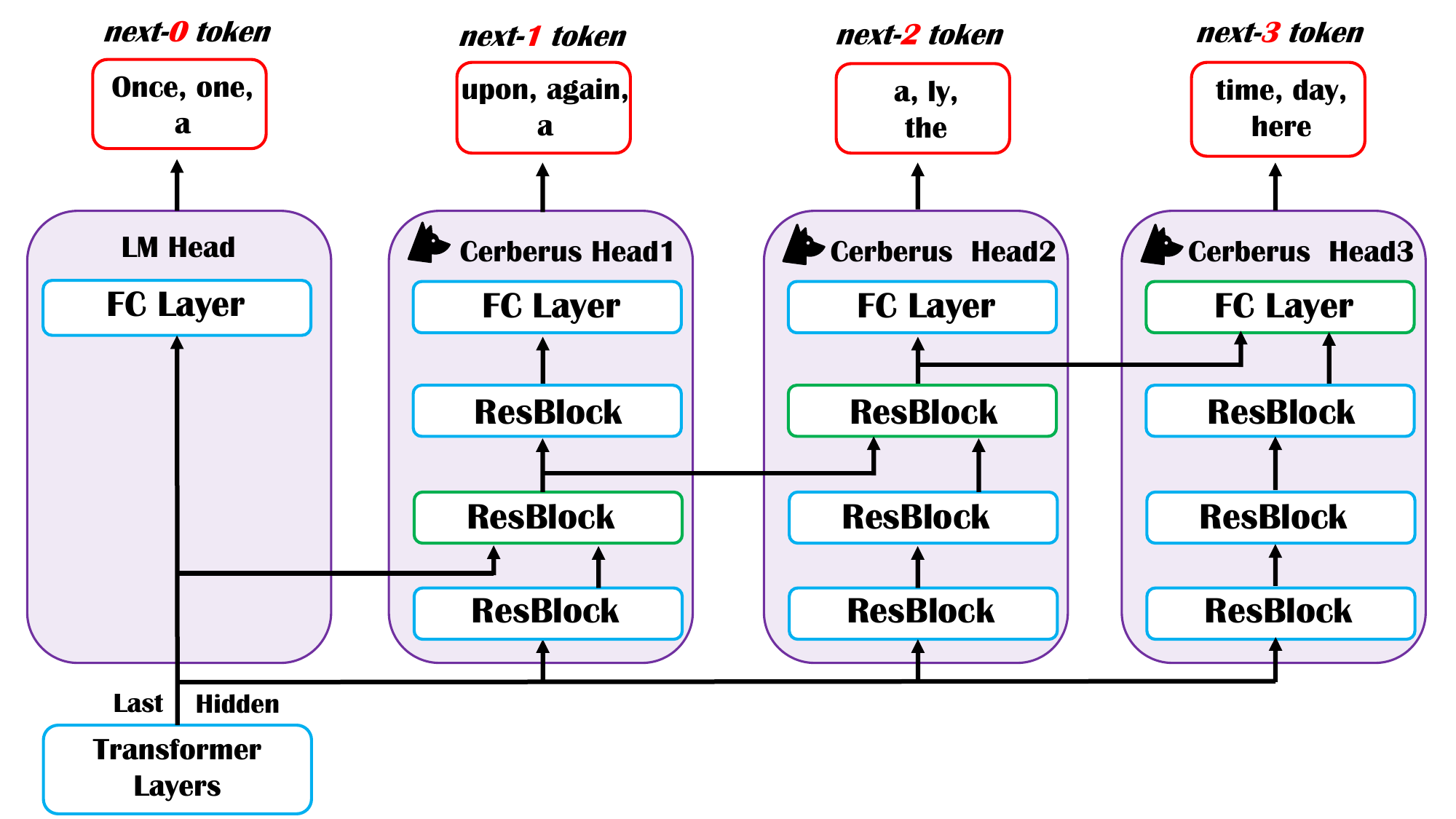}
    \caption{Cerberus heads}
    \label{fig:sub22}
  \end{subfigure}
  \caption{The architecture of Medusa's decoding heads (Medusa heads) and Cerberus's decoding heads (Cerberus heads). Each decoding head is composed of an FC layer and multiple Resblocks, the FC layer refers to a fully connected layer, and the detail of the Resblock is presented in Section \ref{sec:cerberus head} and Section \ref{sec:gating mechanism}.}
\end{figure*}

In this section, We will elaborate on Cerberus and how it effectively addresses the issues in existing parallel decoding frameworks. Cerberus introduces two following modifications. (a) Cerberus heads, a novel decoding head paradigm that introduces sequential knowledge within different decoding heads while maintaining the parallelism of execution. (b) An entropy-based gating mechanism allows LLM to adaptively choose different decoding approaches during each decoding step.

\subsection{Cerberus Heads}
\label{sec:cerberus head}
The existing paradigm of decoding heads can't balance prediction accuracy and parallelism of execution. 
In Medusa \citep{cai2024medusa}, the decoding heads are referred to as Medusa heads. As illustrated in Figure \ref{fig:sub11}, each Medusa head consists of a fully connected (FC) layer and multiple residual blocks (Resblocks). And each Resblock specifically consists of an FC layer and a residual connection \citep{he2016deep}. Given the hidden states \textit{h\textsubscript{i}} for the \textit{i}-th head, the computation flow of the Resblock can be defined as follows:

$$
\text{Output} = \sigma_{\text{SiLU}} \left( (W \cdot h_i + b) + h_i \right)
$$

During inference, each Medusa head independently predicts tokens at different positions while using the same state information from the last transformer layers of LLM. Although this independence enables the heads to generate tokens in parallel, it reduces the prediction accuracy.

To improve the prediction accuracy, other researchers (\citealp{ankner2024hydra}; \citealp{zhang2024recurrent}) introduce a serial connection between the entire decoding heads. Although this approach can help the decoding heads capture more contextual information and improve the prediction accuracy, it will lose the parallelism of the decoding heads in this case, since each head needs to wait for the previous one to complete generation.

Considering all these issues, we introduce the Cerberus Heads (Figure \ref{fig:sub22}). To maintain the simplicity of the architecture, we employed the same construction as Medusa heads, where each head still consists solely of Resblocks and a single FC layer. 

To strengthen the association of each head and improve the prediction accuracy, we still attempt to introduce a sequential connection between the decoding heads. However, we choose to add sequential connections between the Resblocks within the decoding head instead of simply introducing a serial connection between all the decoding heads.

To achieve this, we create a special Resblock (depicted as a green square in Figure \ref{fig:sub22}) within the Cerberus head to receive two hidden states. Specifically, in the $i$-th decoding head, given the hidden states \textit{h\textsubscript{i}} and the hidden states \textit{h\textsubscript{i-1}} from the \textit{i-1}-th head's Resblock, the computation flow of the special Resblock can be formulated as follows:

$$
h = \text{Concat}\left(h_i, h_{i-1}\right)
$$
$$
\text{Output} = \sigma_{\text{SiLU}} \left( (W \cdot h + b) + h \right)
$$

Each decoding head is now equipped to receive information not only from the last hidden states but also from the prior decoding heads, by adding this sequential connection to introduce sequential knowledge. Furthermore, since our implementation is inside the decoding heads rather than outside the entire heads, each decoding head can start decoding simultaneously to maintain the parallelism of decoding heads.

\subsection{Entropy-based Gating Mechanism}
\label{sec:gating mechanism}

Based on our first key observation (Section \ref{unncessary overhead}): During the entire decoding process, some simple decoding steps can be effectively solved using parallel decoding, while some challenging decoding steps that confuse the LLM are only suitable for auto-regressive decoding. Our goal is for the LLM to adaptively choose the appropriate decoding approach at each step.

\begin{algorithm}
\caption{Workflow of each decoding step in Cerberus}
\label{alg:algorithm1}
\begin{algorithmic}[1]
\Statex \textbf{Require:} the number of hidden layers $N$, the transformer layer $TL$, the LLM’s head $LH$, the Cerberus head $CH$, the threshold for entropy $T$, input sequence $x$
\State $h \gets \text{embedding}(x)$
\For{$i$ in range $1,\ldots,N$}
    \State $h \gets TL_i(h)$
\EndFor
\State $h\_last \gets h$
\State $P \gets \text{Softmax}(h\_last)$
\State $S \gets \text{Entropy}(P)$
\If{$S > T$}
    \Statex \textcolor{blue}{\ \ \ \ \ \ // Using auto-regressive decoding}
    \State $O \gets LH(h\_last)$
\Else
    \Statex \textcolor{blue}{\ \ \ \ \ \ // Using parallel decoding}
    \State $Candidates \gets (LH[h\_last],$
    \State $(CH_0[h\_last]), \ldots, (CH_i[h\_last]))$
    \State $O \gets Tree\_Verify (Candidates)$
\EndIf
\State $Output \gets O$
\State \Return $Output$
\end{algorithmic}
\end{algorithm}

To achieve this, we employ the gating mechanism to select whether to use auto-regressive decoding or parallel decoding during each decoding step, inspired by the Mixture-of-Experts (MOE) (\citealp{jacobs1991adaptive}, \citealp{shazeer2017outrageously}), which also uses a gating network to select different sub-models to cope with different tasks.

Specifically, We decide to use an entropy-based gating mechanism, since we have observed that the entropy of the last hidden states can reflect the prediction accuracy of LLMs and help us assess the model's confidence level in the current prediction (Section \ref{entropy}). The higher the entropy value, the less the final prediction accuracy, and the less confident the model is in the current prediction. Given the last hidden states \textit{h}, the calculation process of entropy can be defined as follows:

$$
P = \text{Softmax}\left(h\right)
$$
$$
entropy(P) = -\sum_{i=1}^{K} P(x_i) \log_2 P(x_i)
$$

This gating mechanism can be easily integrated into the LLM. For any model, we only have to simply add an operator to calculate entropy after the last hidden layer of the LLM and set a threshold for comparison. If the entropy value is greater than this threshold, it indicates that the parallel decoding is not a suitable solution in this case, hence only auto-regressive decoding will be used to finish this generation. By setting this gating mechanism, the workflow of each decoding step in Cerberus is presented in Algorithm \ref{alg:algorithm1}.

Unlike existing parallel decoding frameworks, which implement parallel decoding for the entire decoding process. Our framework Cerberus can adaptively decide whether to use auto-regressive decoding or parallel decoding based on the model's confidence level during each decoding step. Only by implementing parallel decoding at the decoding steps is the LLM confident in reducing unnecessary overheads caused by parallel decoding.

\section{Experiments}

In this section, we conduct several experiments to evaluate the Cerberus on various downstream tasks and make a comparison with the Medusa \citep{cai2024medusa} and auto-regressive decoding.

\subsection{Experimental Setup}
\label{experimental setup}

\textbf{Models} Both the Cerberus and Medusa will be employed on the Vicuna-7B \citep{chiang2023vicuna}, which is fine-tuned from the LLaMA \citep{touvron2023llama} model.\\

\begin{table*}
  \centering
  \begin{tabular}{c|c|ccccccccc}
    \hline
    \textbf{Metrics} & \textbf{Approach} &  \textbf{Avg.} & \textbf{WT} & \textbf{RP} & \textbf{ET} & \textbf{RS} & \textbf{MA} & \textbf{CD} & \textbf{ST} & \textbf{HT}\\
    \hline
    \multirow{3}{*}{\centering Tokens/Second} & Cerberus & \textbf{80.35} & \textbf{81.61} & \textbf{74.87} & \textbf{79.06} &\textbf{85.48} &\textbf{64.95} &\textbf{75.98} &\textbf{90.61} &\textbf{90.22} \\ \cline{2-2}
     & Medusa & 77.21 & 78.68 & 67.30 &77.21 & 81.47 &62.30 &73.06 &90.10 &87.57                         \\ \cline{2-2}
     & Vanilla & 44.17 & 45.17 & 42.22 & 44.61 & 45.78 & 40.93 & 45.15 & 44.67 & 44.83                         \\ \hline
    \multirow{3}{*}{\centering Quality} & Cerberus & 5.90 & 7.08 & 7.25 &4.65 & 2.65 &3.15 &\textbf{5.68} &\textbf{7.80} &\textbf{8.90}                           \\ \cline{2-2}
     & Medusa & 5.84 & 7.15 & 7.30 &4.85 & 2.60 &\textbf{3.30} &5.65 &7.45 &8.45                           \\ \cline{2-2}
     & Vanilla & \textbf{5.92} & \textbf{7.35} & \textbf{7.65} & \textbf{4.90} & \textbf{2.75} & \textbf{3.30}& 5.45 & 7.70 & 8.25                          \\ \hline
  \end{tabular}
  \caption{
   Comparison under all downstream tasks of MT-Bench, with setting tree paths to 120 and the entropy threshold to 0.59. The vanilla refers to Auto-regressive decoding, and Quality is the score obtained by using GPT-4 as the judge in MT-Bench.
  }
  \label{detailed comparison}
\end{table*}

\noindent \textbf{Dataset} We evaluate the Cerberus on the MT-Bench \citep{zheng2024judging}, which is a dataset used to evaluate the ability of LLMs in multi-round conversations and instruction adherence. The MT-Bench is composed of eight different types of questions, including writing (\textbf{WT}), roleplay (\textbf{RP}), extraction (\textbf{ET}), reasoning (\textbf{RS}), math (\textbf{MA}), coding (\textbf{CD}), stem (\textbf{ST}), and humanities (\textbf{HT}). For each category, researchers manually designed 10 multi-round questions.\\

\noindent \textbf{Tree Settings} Both Cerberus and Medusa utilize the tree attention to verify the generated token candidates to find the final tokens to be accepted in parallel. During tree verification, the tree path is namely the path of the verification process, and the number of tree paths is equal to the number of token candidates to be verified. For example, if top-$k$ options are selected for each decoding head, the solution space of the 4 decoding heads will generate a total of $k+k^2+k^3+k^4$ tree paths. To reduce the size of the solution space, some paths can be manually selected for verification, rather than verifying all paths. In the technical report of Medusa, the k is set to 10, hence the total number of verified paths should be $10+10^2+10^3+10^4$, but the actual number of verified paths (\textit{i.e.}, tree paths) was manually set to 63. For more robust experimental results, we set the $k$ to 10 aligned with the Medusa and conducted experiments under three tree path settings (\textit{i.e.}, 63, 120, 150) rather than only one tree setting.\\

\noindent \textbf{Training} Both the Cerberus and Medusa are trained while freezing all weights of the original LLM, only weights of the decoding heads will change during the training process. We set up 4 decoding heads for Cerberus and Medusa, each decoding head consists of four Resblocks and one FC layer, with identical initial weights following the configuration in the Medusa \citep{cai2024medusa}. We also set the learning rate to 4e-4 to implement training. In this configuration, both Cerberus and Medusa require training with 2x NVIDIA A100 GPU for about 3 hours.

\subsection{End-to-End Result}

Table \ref{detailed comparison} shows the detailed comparison among three decoding approaches when the number of tree paths is set to 120. Cerberus achieves the best inference speed in all downstream tasks of MT-Bench. When regarding the auto-regressive decoding as the baseline, the average acceleration obtained by Cerberus is 10\% higher than that of Medusa. In the tasks of Roleplay, Cerberus’s acceleration increment can even achieve 30\% higher than Medusa's. Furthermore, in terms of generation quality, Cerberus also performs better than Medusa, with only a slight decrease compared to auto-regressive decoding. Overall, Cerberus shows great performance in both inference speed and generation quality.

\begin{figure*}
  \includegraphics[height=8.5cm,width=\textwidth]{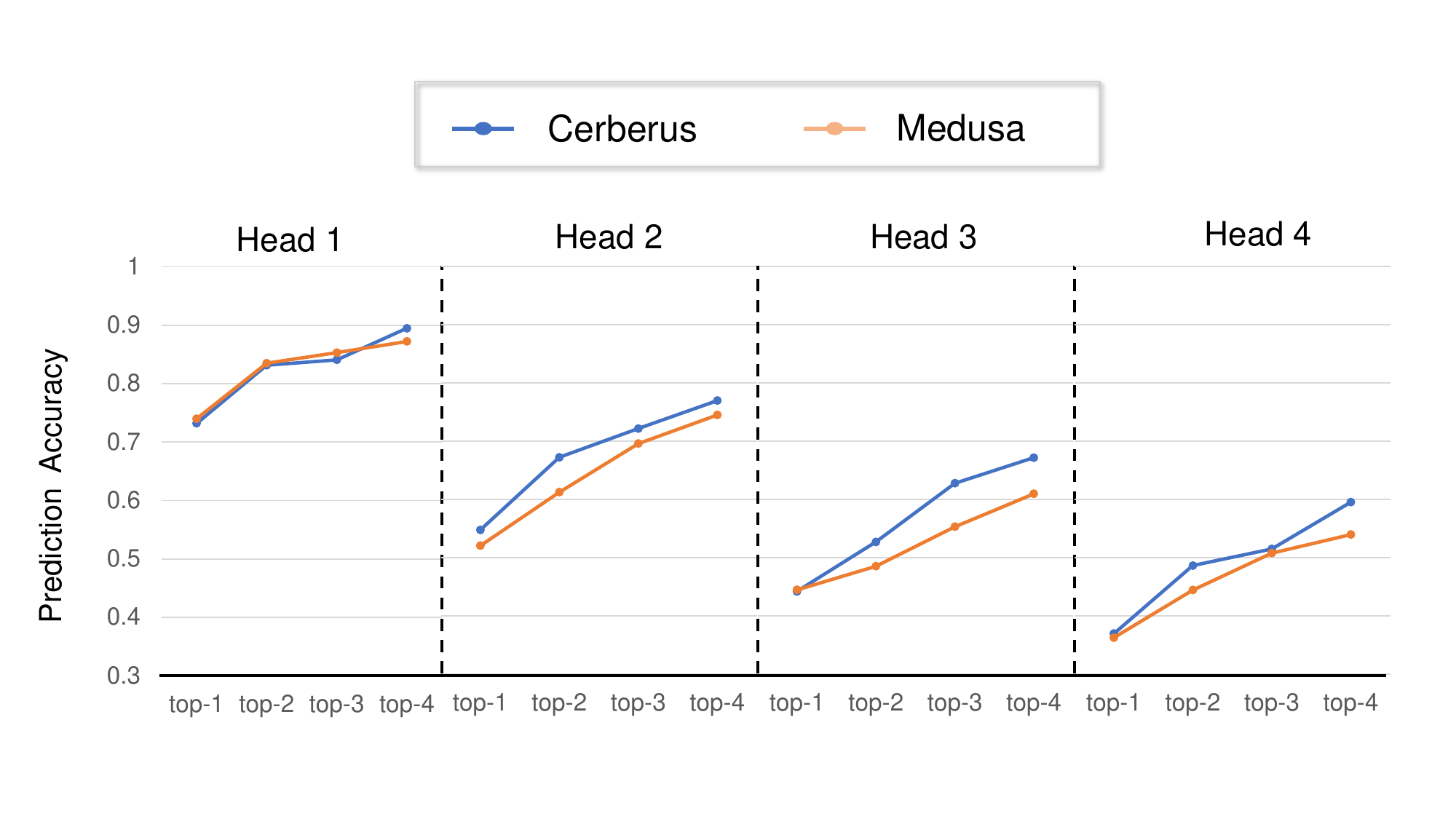}
  \caption{The comparison of top-$k$ accuracy between Medusa heads and Cerberus heads.}
  \label{fig:prediction accuracy}
\end{figure*} 

\begin{figure}[H]
  \includegraphics[width=\columnwidth]{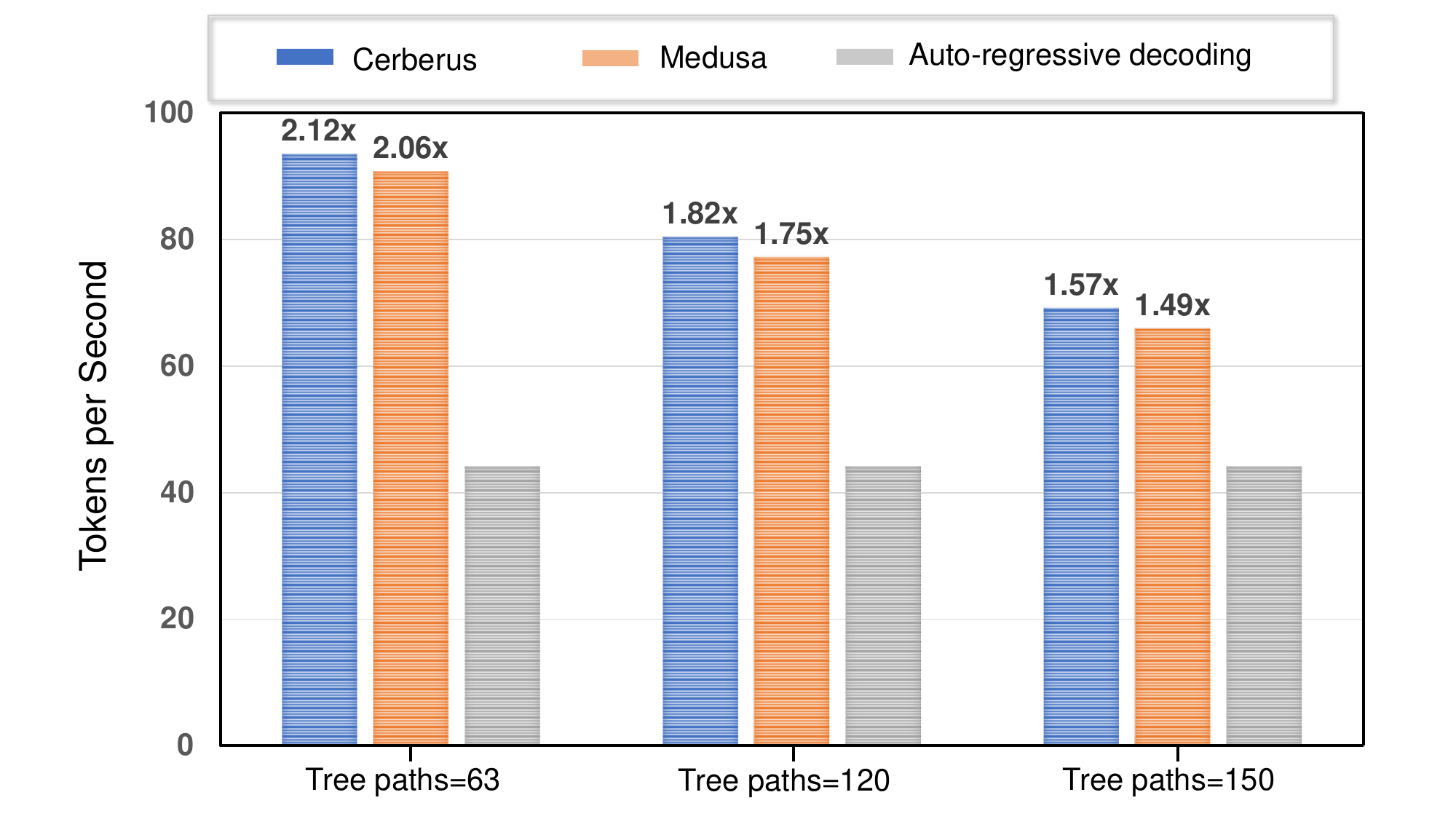}
  \caption{Speedup ratio of Cerberus and Medusa under three different tree settings. Cerberus achieves the best acceleration in all settings.}
  \label{fig:tokens per second}
\end{figure}

To demonstrate the robustness of Cerberus's performance, we then compare the inference speed of Cerberus, Medusa, and auto-regressive decoding under three types of tree settings. As shown in Figure \ref{fig:tokens per second}, Cerberus can achieve a 2.12x speedup compared to auto-regressive decoding, surpassing Medusa's 2.06x speedup, when aligned with the original setting of Medusa's technical report \citep{cai2024medusa} (\textit{i.e.}, when the number of tree paths is set to 63). Moreover, Cerberus still achieves the best inference speed for the other two tree settings, which can demonstrate Cerberus's robustness in accelerating inference.

\section{Ablation Study}
\label{sec:x}
To obtain a deeper understanding of our approach, we independently conduct a series of ablation experiments on each component of Cerberus.\\

\noindent \textbf{Ablation on Gating Mechanism} To evaluate the effectiveness of the gating mechanism separately, we compared the inference speed of Cerberus and Cerberus without the gating mechanism (Cerberus w/o gating) on MT-Bench. As shown in Table \ref{detailed comparison of gating}, Cerberus performs better than Cerberus w/o gating in all tasks of the MT-Bench. This suggests that the gating mechanism plays an important role in Cerberus.\\

\begin{table}[H]
  \centering
  \begin{tabular}{c|cc}
    \hline
    \multirow{2}{*}{\textbf{Task}} & \multicolumn{2}{c}{\textbf{Tokens/Second}} \\
    & \textbf{Cerberus} & \textbf{Cerberus w/o gating} \\
    \hline
    \textbf{Avg.} &80.16 & 78.26 \\
    \hline
    \textbf{WT} &81.41 & 78.16 \\
    \hline
    \textbf{RP} &74.67 & 71.20 \\
    \hline
    \textbf{EC} &78.31 & 76.64\\
    \hline
    \textbf{RS} &85.18 & 81.01\\
    \hline
    \textbf{MA} &64.95 & 63.79\\
    \hline
    \textbf{CD} &75.99 &  75.09\\
    \hline
    \textbf{ST} &90.54 & 90.26\\
    \hline
    \textbf{HM} &90.23 & 89.91\\
    \hline
  \end{tabular}
  \caption{
   Comparison between Cerberus and Cerberus without using the gating mechanism, tested on eight types of downstream tasks of the MT-Bench.}
  \label{detailed comparison of gating}
\end{table}

\noindent \textbf{Ablation on Decoding Heads} To evaluate the performance of Cerberus heads separately, we compare the top-$k$ prediction accuracy of Cerberus heads and Medusa heads, which can well reflect the decoding head's ability to predict tokens correctly.
As shown in Figure \ref{fig:prediction accuracy}, the overall performance of the Medusa heads in top-$k$ accuracy is inferior to that of Cerberus heads. Especially for the prediction of the rear positions like the head3 and head4, the Cerberus heads can show more obvious advantages since the prediction gets harder. This indicates that Cerberus heads capture longer contexts and have better performance than the Medusa heads by introducing sequential knowledge between decoding heads.\\

\section{Conclusion}
\label{sec:bibtex}
In this paper, we propose Cerberus, an adaptive and effective parallel decoding framework that can be seamlessly integrated into existing LLMs. It incorporates two key components: (1) a novel paradigm for decoding heads that introduces sequential knowledge without compromising execution parallelism, and (2) an entropy-based gating mechanism that enables LLMs to select the most appropriate decoding approach at each step. In experiments on MT-Bench, Cerberus achieves up to a 2.12x speedup compared to auto-regressive decoding and outperforms Medusa in both acceleration and generation quality.

\section*{Limitations}
Although our work provides an efficient parallel decoding framework, there are still several limitations as follows:

• Cerberus heads introduce sequential knowledge while also introducing more parameters, which requires more computation resources.

• For different LLMs and tasks, different entropy thresholds may be required, hence finding a suitable entropy threshold can be challenging. However, we still rely on manual experimentation to determine the optimal entropy threshold so far. How to adaptively determine a suitable entropy threshold for different LLMs and tasks is still an issue that remains an area for further research.

\bibliography{custom}

\end{document}